%% file: root.tex
\documentclass[a4paper,conference]{IEEEtran}
\usepackage[pdftex]{graphicx}
\DeclareGraphicsExtensions{.pdf}

\usepackage{booktabs}
\usepackage{textcomp}

\begin{document}

\title{Self-supervised Learning for \\Astronomical Image Classification}

\author{\IEEEauthorblockN{Ana Martinazzo, Mateus Espadoto, Nina S. T. Hirata}
\IEEEauthorblockA{Institute of Mathematics and Statistics\\
University of S\~{a}o Paulo\\
S\~{a}o Paulo, Brazil\\
Email: {amartina $\vert$ mespadot $\vert$ nina}@ime.usp.br}}

\maketitle

\begin{abstract}
In Astronomy, a huge amount of image data is generated daily by photometric surveys, which scan the sky to collect data from stars, galaxies and other celestial objects. In this paper, we propose a technique to leverage unlabeled astronomical images to pre-train deep convolutional neural networks, in order to learn a domain-specific feature extractor which improves the results of machine learning techniques in setups with small amounts of labeled data available. We show that our technique produces results which are in many cases better than using ImageNet pre-training.
\end{abstract}


%
\IEEEpeerreviewmaketitle

\section{Introduction}
\label{sec:intro}
\input{010_intro.tex}

\section{Background}
\label{sec:related}
\input{020_background.tex}

\section{Method}
\label{sec:method}
\input{030_method.tex}

\section{Results}
\label{sec:results}
\input{040_results.tex}

\section{Discussion}
\label{sec:discussion}
\input{050_discussion.tex}

\section{Conclusion}
\label{sec:conclusion}
\input{060_conclusion.tex}

\section*{Acknowledgment}

\noindent This study was financed in part by FAPESP (2015/22308-2, 2017/25835-9 and 2018/25671-9) and the Coordena\c{c}\~{a}o de Aperfei\c{c}oamento de Pessoal de N\'{i}vel Superior - Brasil (CAPES) - Finance Code 001.

\IEEEtriggeratref{16}
\IEEEtriggercmd{\enlargethispage{-6.5in}}

\bibliographystyle{IEEEtran}
\bibliography{refs}

\end{document}

%% file: 010_intro.tex
Modern Astronomy relies on photometric surveys to gather data from objects in the sky. In contrast with earlier techniques, where data was collected from a few objects at a time and with great detail, using spectroscopy, photometric surveys enable the collection of vast amounts of image data at once, thus demanding the use of automated data analysis techniques, such as machine learning (ML).

The usage of ML techniques in Astronomy is not new, but was mostly focused on processing of catalog data~\cite{moore2006,gauci2010machine,ball2010data,ivezic2014statistics}, which is based on features extracted from photometric data and stored in table format. In the past few years, works using Deep Learning~\cite{bengio2012deep,krizhevsky2012imagenet} techniques applied to images started to appear, trying to address problems in star/galaxy classification~\cite{kim2016star}, galaxy morphology~\cite{dieleman2015rotation} and merging galaxies detection~\cite{ackerman2018}, and more recently, a systematic comparison between different Deep Learning models and Transfer Learning~\cite{pan2009survey} setups applied to different problems was done as well~\cite{VISAPP}.

As we see it, researchers are only starting to use Deep Learning techniques in Astronomy, which might have to do with the fact that those techniques usually require large amounts of labeled data to work properly, and that despite the availability of a huge amount of data in astronomy, typically only a small percentage is labeled for specific problems. Labeling astronomical data is costly, as it requires expert knowledge or crowdsourcing efforts, such as Galaxy Zoo~\cite{gzoo2008}.

In this work, we propose the usage of \emph{self-supervised} learning for astronomical image data, where a large neural network is pre-trained with large amounts of unlabeled data as input and astronomical properties as output, which can be cheaply computed based on the input images alone, in a regression setup. We show that with this pre-training, accuracy is improved for downstream classification tasks even with very small amounts of labeled training data available, and that in most cases the results are better than the ones obtained by doing Transfer Learning based on ImageNet pre-training~\cite{imagenet}, given the same amount of labeled data.

The main contributions of this work are as follows:

\begin{itemize}
    \item we present a method that uses cheaply computed astronomical properties to train a high-quality, domain-specific feature extractor based on unlabeled data;
    \item we train classifiers using many labeled datasets, to demonstrate the technique in different settings;
    \item we show that our technique makes it possible to obtain peak classification performance with less labeled data when compared with ImageNet pre-training.
\end{itemize}

The structure of this paper is as follows. Section~\ref{sec:related} presents background on astronomical data and self-supervised learning. Section~\ref{sec:method} details our experimental setup. Section~\ref{sec:results} presents our results, which are next discussed in Section~\ref{sec:discussion}. Section~\ref{sec:conclusion} concludes the paper.

%% file: 020_background.tex
\subsection{Astronomical data}

Astronomical data can be found in two formats: spectra and images. Spectra consist of nearly continuous measures of energy fluxes emitted or absorbed by an object as a function of wavelength. Their fine-grained resolutions allow reliable categorization of objects by matching observed energy curves to theoretical ones. This information, in turn, is used to validate theories of how and when a given type of object formed, how it evolved and what are the physical processes involved in its formation and evolution. In spite of providing such rich and fine-grained information, spectra are hardly scalable: a spectrograph requires hours of exposure in order to collect enough photons across all wavelengths. Spectral data, thus, is expensive and scarce. There are datasets of spectra that have become significantly large, such as the one from the Sloan Digital Sky Survey (SDSS)~\cite{Abazajian09}, and that can be used as ground-truth for fully supervised tasks. Still, there is an inherent bias in how objects were chosen for spectroscopic observation.

In order to build datasets that are more representative of the universe at large, images are used. Astronomical images can be understood as a discretized version of spectra, in which energy flux measures are grouped in broader ranges of wavelength. Image-based sky surveys yield information about numbers of objects that are orders of magnitude larger than spectroscopic ones. Even though these information usually come at the cost of larger uncertainties, they are vital for defining distributions of objects at large scales, for capturing transient phenomena (such as asteroids and supernova explosions), and also for detecting unusual phenomena (the so-called outliers) that may be more carefully studied with a spectrograph afterwards. Besides, images provide rich morphological information that is usually absent in spectra, and that can be used for studying formation and evolution of objects in different ways.

Astronomical image acquisition is carried out by collecting photons with CCD cameras mounted to telescopes containing a set of filters with different passbands. The resulting images are three-dimensional arrays wherein the depth dimension corresponds to the amount of filters used, and each pixel corresponds to the count of photons that passed through a filter in that position. Tasks such as object segmentation (using traditional, unsupervised computer vision techniques) and estimation of properties of objects are performed over these images. After this, a catalog of objects of unknown classes with position coordinates and estimates of properties is generated. The most used tool for this kind of low-level processing is called SExtractor~\cite{Bertin96}.

Raw astronomical images usually come in more than three channels, making them impossible to visualize using the standard color model for digital images, RGB. However, there are algorithms~\cite{Lupton03} that combine information from multiple channels, enhancing relevant details from each passband, and generate RGB composite images. These are mostly used for scientific communication, but are now also used for research at the interface between Deep Learning and Astronomy. In this work, we use RGB composite images made available by various sky surveys.

\subsection{Self-supervised Learning}

Self-supervised learning consists of using unlabeled data and some attribute that can be easily and automatically generated from these data for training models on a pretext task, with the objective of learning representations that are useful for other tasks.
Some methods of producing pretext tasks for image data are:

\begin{itemize}
	\item Clustering~\cite{Caron18}: features extracted from a model are clustered and their cluster assignments are used as pseudo-labels for iteratively training the model.
	\item Image Rotation~\cite{Gidaris18}: four copies of images are generated by rotating them by $0^{\circ}$, $90^{\circ}$, $180^{\circ}$ and $270^{\circ}$, and a model is trained to predict which rotation was applied to each image.
	\item Relative Patch Location~\cite{Doersch15}: pairs of patches from images are generated, and a model is trained to predict the position of the second patch relative to the first.
\end{itemize}

In Astronomy, most data is unlabeled, but useful properties of the objects, such as brightness and size, can be readily computed from analytical models, without supervision. This yields an ideal setting for using self-supervision, where computed properties are used as targets for regression tasks. In this way, unlabeled data are included in the learning process, leading to representations that are closer to the distribution of the observed objects in the universe at large.

Astronomical properties alone are commonly used as input features for object classification~\cite{Sesar17, Costa-Duarte19}. However, we believe that combining images with computable properties of astronomical objects into a single representation through self-supervised learning leads to more powerful, representative features that can achieve better performance on more difficult tasks, such as outlier detection or clustering.


%% file: 030_method.tex
\subsection{Pretext task}
\label{sec:method_pretext}

A pretext task is a task in which known attributes of each data point are used as intermediate targets for training models in a supervised manner, with the objective of learning representations that can be used in higher-level tasks such as classification or clustering. We choose the prediction of magnitudes from unlabeled images of astronomical objects as our pretext task.

Magnitude is a dimensionless measure of the brightness of an object in a given passband. Magnitudes are defined in a logarithmic scale, such that a difference of 1 in magnitude corresponds to a difference of $100^{1/5}$ in brightness, and magnitude values found in modern sky surveys are roughly inside the range [0, 30], where lower values correspond to brighter objects. To ease training and convergence, we rescale magnitudes by dividing them by 30. Magnitudes are computed with preprocessing tools such as SExtractor~\cite{Bertin96} and made available by the sky survey teams for the scientific community.

We use a subset of the first data release of the Southern Photometric Local Universe Survey (S-PLUS)~\cite{Oliveira19}, a sky survey aimed at imaging the Southern sky in twelve filters. There are, therefore, twelve magnitudes per object to be used as targets. The first S-PLUS data release includes both image cutouts and a catalog of detected objects, their estimated properties and their corresponding uncertainties. Estimates of magnitudes in the catalog have a precision of $0.01$ and come with uncertainty estimates.

Table~\ref{table:magnitudes} shows magnitude values and their uncertainties for a randomly sampled object. Rows are ordered by passbands from smallest (u) to largest (z) wavelengths. The filters u, g, r, i and z are a widely used set of broad filters, known as the \emph{ugriz} system. They refer to ultraviolet, green, red, infrared, and near-infrared. The other filters refer to narrow filters whose objective is to capture information about phenomena that happen at specific wavelengths. It can be seen that uncertainties vary significantly per filter.

\begin{table}[!ht]
\centering
\caption{An example of magnitude values \newline used as targets for the pretext task}
\label{table:magnitudes}
\begin{tabular}{lc}
\toprule
u		& 19.87 \textpm 0.04 \\
f378    & 19.93 \textpm 0.06 \\
f395    & 19.95 \textpm 0.09 \\
f410    & 19.42 \textpm 0.06 \\
f430    & 19.34 \textpm 0.05 \\
g       & 19.16 \textpm 0.02 \\
f515    & 19.09 \textpm 0.04 \\
r       & 18.96 \textpm 0.02 \\
f660    & 18.93 \textpm 0.02 \\
i       & 18.82 \textpm 0.02 \\
f861    & 18.78 \textpm 0.03 \\
z       & 18.80 \textpm 0.03 \\
\toprule
\end{tabular}
\end{table}

We filter out every object for which any of the magnitudes has an uncertainty higher than $0.1$. This uncertainty threshold keeps a balance between having a dataset of reasonable size and removing examples with noisy targets. We also filter out all labeled objects, which will later be used for downstream tasks, in order to avoid leaking from the pretext task to downstream tasks.

The resulting unlabeled set contains 205321 objects, which are split into 80\% for training, 10\% for validation and 10\% for testing. Results of the pretext training are described in Section~\ref{sec:results_pretext}.

\subsection{Downstream tasks}
\label{sec:method_downstream}

Six datasets with varying degrees of difficulty are selected for downstream classification tasks. Two of them are extracted from the Southern Photometric Local Universe Survey (S-PLUS)~\cite{Oliveira19}, the sky survey also used in the pretext task, and the other four from the Sloan Digital Sky Survey (SDSS)~\cite{Abazajian09}, a sky survey that sweeps the Northern sky in five filters. Each dataset is briefly described below.

\vspace{5pt}

\noindent \textbf{Star/Galaxy (SG):} 50090 images divided into two classes: \emph{Stars} (27981) and \emph{Galaxies} (22109), extracted from Data Release 1 of S-PLUS. This dataset is considered the easiest among the six, since it has a reasonable amount of data, classes are balanced, and it contains easily identifiable examples of stars and galaxies.

\begin{figure}[htb!]
    \centering
        \includegraphics[width=0.47\textwidth]{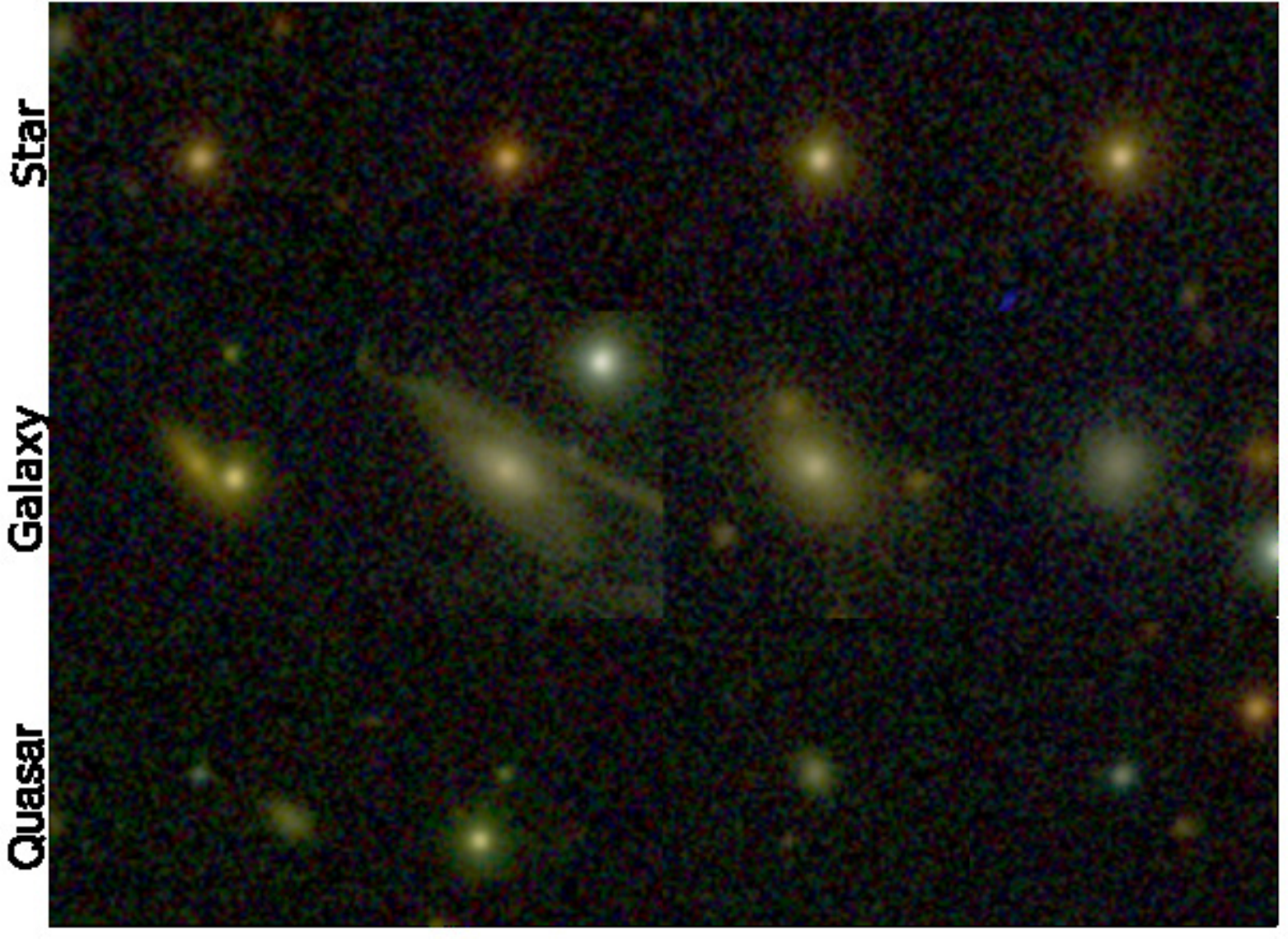}
    \caption{Sample images from the SGQ dataset.}
    \label{fig:sgq}
\vspace{-0.15cm}
\end{figure}

\medskip
\noindent \textbf{Star/Galaxy/Quasar (SGQ):} 54000 images divided into three balanced classes: \emph{Stars} (18000), \emph{Galaxies} (18000), and \emph{Quasars} (18000), also extracted from Data Release 1 of S-PLUS. This dataset is a harder variant of the \emph{SG} dataset, since it contains fainter examples of galaxies and stars, and also contains quasars (quasi-stellar radio sources). Quasars are small objects that are easily confused with stars. See examples in Fig.~\ref{fig:sgq}.

\begin{figure}[htb!]
    \centering
        \includegraphics[width=0.47\textwidth]{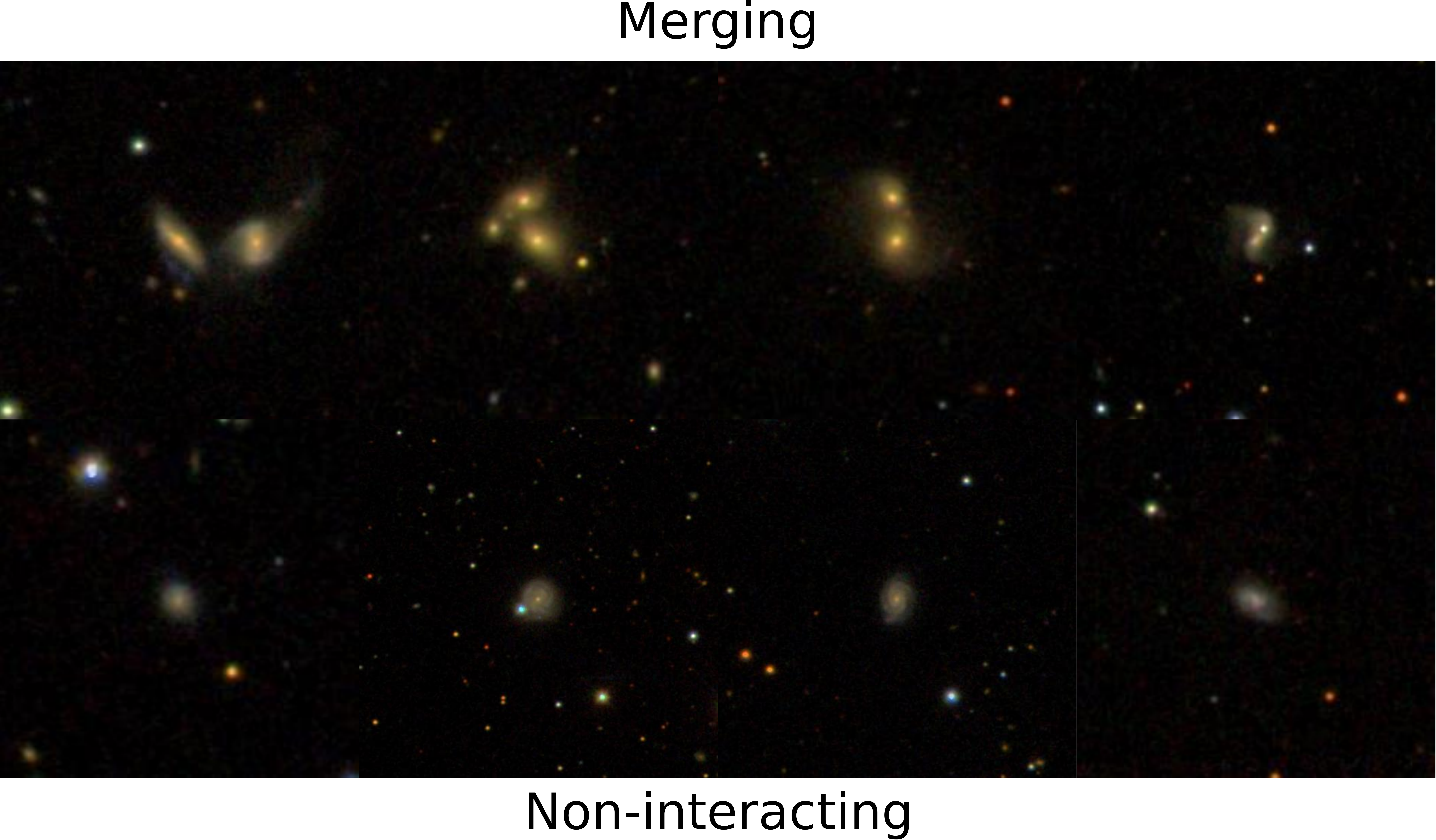}
    \caption{Sample images from the MG dataset.}
    \label{fig:mg}
\vspace{-0.15cm}
\end{figure}

\medskip\noindent \textbf{Merging Galaxies (MG):} 15766 images divided into two classes: \emph{Merging} (5778) and \emph{Non-interacting} (9988) galaxies, extracted from Data Release 7 of SDSS. This dataset is sufficiently large, but its instances are not as clearly separable as stars and galaxies. See examples in Fig.~\ref{fig:mg}.

\medskip\noindent \textbf{Galaxy Morphology, 2-class (EF-2):} 3604 images divided into two classes: \emph{Elliptical} (289) and \emph{Spiral} (3315) galaxies, extracted from the EFIGI~\cite{baillard2011efigi} dataset, which is based on SDSS. This dataset is highly unbalanced towards images of \emph{Spiral} galaxies.

\medskip\noindent \textbf{Galaxy Morphology, 4-class (EF-4):} 4389 images divided into four classes: \emph{Elliptical} (289), \emph{Spiral} (3315), \emph{Lenticular} (537) and \emph{Irregular} (248) galaxies, extracted from the EFIGI dataset. The additional classes make the classification problem harder, since the objects are not as clearly identifiable as in the 2-class subset and classes are highly unbalanced as well.

\begin{figure}[htb!]
    \centering
        \includegraphics[width=0.47\textwidth]{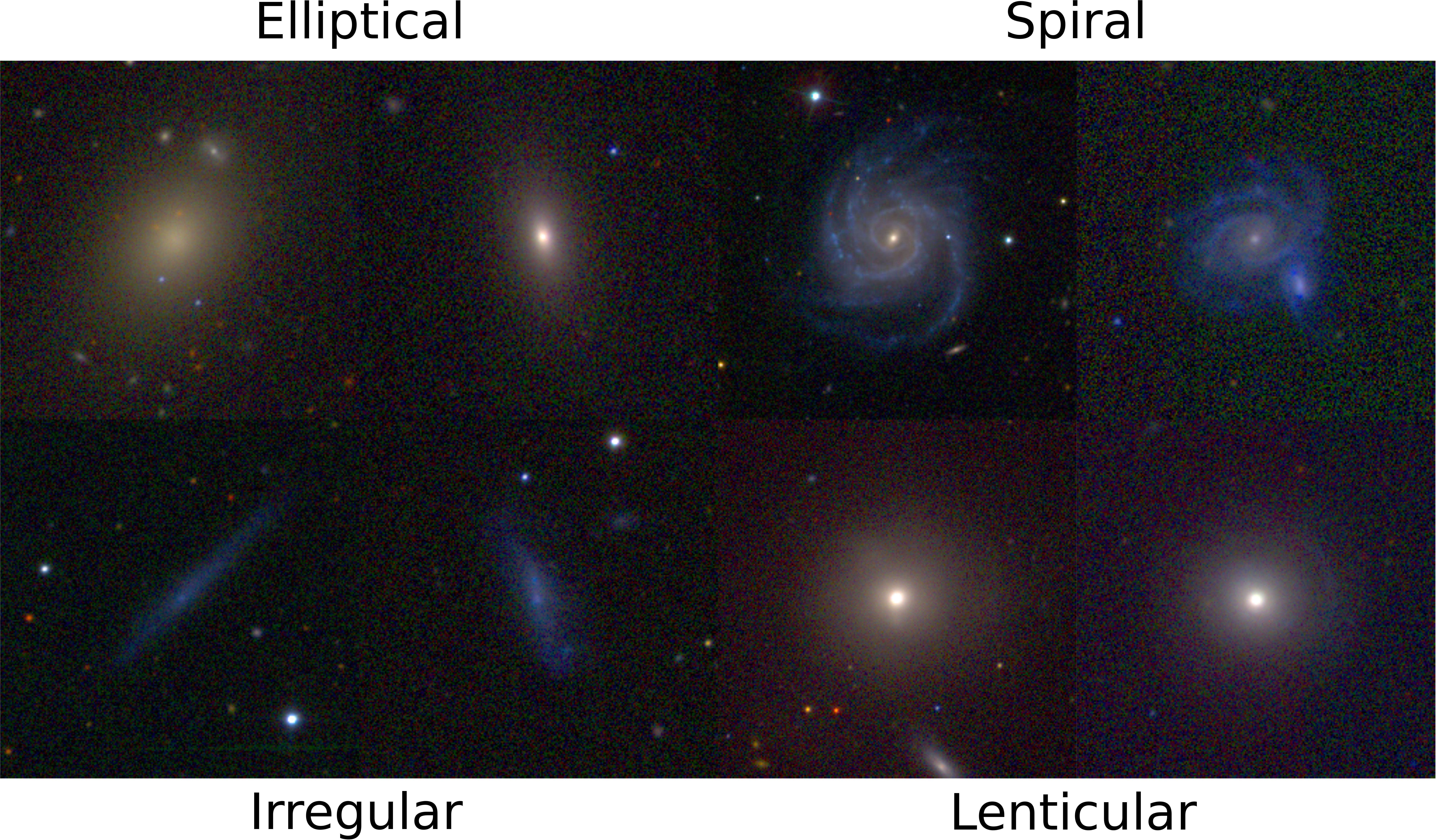}
    \caption{Sample images from the EF-2, EF-4 and EF-15 datasets.}
    \label{fig:ef}
\vspace{-0.15cm}
\end{figure}

\medskip\noindent \textbf{Galaxy Morphology, 15-class (EF-15):} 4327 images divided into fifteen classes: \emph{Elliptical:-5} (227), \emph{Spiral:0} (196), \emph{Spiral:1} (257), \emph{Spiral:2} (219), \emph{Spiral:3} (517), \emph{Spiral:4} (472), \emph{Spiral:5} (303), \emph{Spiral:6} (448), \emph{Spiral:7} (285), \emph{Spiral:8} (355), \emph{Spiral:9} (263), \emph{Lenticular:-3} (189), \emph{Lenticular:-2} (196), \emph{Lenticular:-1} (152), and {Irregular:10} (248) galaxies, also extracted from the EFIGI dataset. The numbers after the names come from the Hubble Sequence~\cite{hubble1982realm}, which is a standard taxonomy used in astronomy for galaxy morphology. This is the hardest dataset among the evaluated, since it has only a few hundred observations for each class. See examples for EF-2, EF-4 and EF-15 in Fig.~\ref{fig:ef}.

\vspace{5pt}

Labels for the S-PLUS datasets were obtained by matching astronomical coordinates of the objects between the S-PLUS image catalog and the SDSS spectra catalog, which contains orders of magnitude less samples. Because of that, even though the image catalog may include millions of detected objects, only thousands of them can be reliably labeled and used for supervised training.

\subsection{Training and evaluation setup}
\label{sec:method_training}

In~\cite{VISAPP}, we extensively analyzed training setups for a variety of astronomical object classification tasks. Five CNN architectures, three optimizers, two values of L2 regularization ($\lambda$)~\cite{krogh1992simple} and two values of max-norm constraint ($\gamma$)~\cite{srebro2005rank} were considered. We found that VGG16~\cite{Simonyan14} yielded the best results for all tasks in this context, both when training from scratch and when fine-tuning based on ImageNet weights, and that ImageNet was beneficial in most cases. We also found that VGG16 combined with $\lambda=0$ and $\gamma=2$ yielded the best results for most tasks.

Thus, in this work, we use a VGG16 backbone with $\lambda=0$ and $\gamma=2$ for all tasks. As in~\cite{VISAPP}, we add a 2048-unit fully-connected layer with max-norm constraint, followed by a Dropout layer~\cite{srivastava2014dropout} with $0.5$ probability of dropping units, followed by a final fully-connected layer with $n$ units, where $n$ is either the number of magnitudes or the number of classes. The output of the last layer is passed either through a softmax function, if the task is classification, or through a modified ReLU function that saturates to 1 for $x \ge 1$, if the task is regression. As mentioned earlier, regression targets are divided by 30. For the classification task, cross entropy was used as the loss function, while for the regression task, mean absolute error (MAE) was used.

For the pretext (regression) task, SGD with an initial learning rate of $10^{-3}$ is used. For downstream (classification) tasks, we consider five schemes for comparison:

\begin{itemize}
	\item training from scratch;
	\item extracting features from a model pre-trained on ImageNet;
	\item extracting features from a model pre-trained on the pretext task;
	\item fine-tuning a model pre-trained on ImageNet;
	\item fine-tuning a model pre-trained on the pretext task.
\end{itemize}

When training from scratch, all layers are trained for up to 200 epochs using Adam. When extracting features, all the convolutional layers are kept frozen and the top layers are trained for up to 100 epochs using Adam. When fine-tuning, first all the convolutional layers are frozen and only the top layers are trained for 10 epochs using Adam, then the convolutional layers are unfrozen and trained along with the top layers for up to 200 epochs using SGD with a learning rate of $10^{-4}$. In all cases, training is automatically stopped if validation loss diverges from training loss for more than 10 epochs. Each combination of dataset and scheme is run three times and the average and the standard deviation of the results is reported.

\subsection{Low-data regime}
\label{sec:method_low_data}

We carry out an experiment in which we observe how performance varies with the size of the training set. In order to do that, we start out with a small subset of 100 samples of the training data, and train models with incrementally larger subsets of data until the entire training set is used. From 100 to 1000 samples, an incremental step of 100 is used, to account for variances in very low-data regime. From 1000 to 3000, a step of 500 is used. Finally, from 10000 to 40000, a step of 10000 is used. This yields a total of 18 training runs per dataset and per scheme. The validation set is fixed.

Training from scratch was carried out using SGD with a learning rate of $10^{-4}$, which we empirically found to be more stable than Adam when using little data. This experiment is run only for the S-PLUS datasets, SG and SGQ.

%% file: 040_results.tex
\subsection{Pretext task}
\label{sec:results_pretext}
Training was carried out using the dataset described in Section~\ref{sec:method_pretext}, with architectures and parameters as specified in Section~\ref{sec:method_training}. A MAE of $0.0034$ was achieved on the validation set for the prediction of magnitudes in the [0,1] range. This corresponds to a MAE of $0.1$ on the original magnitude scale. Given that the magnitudes have uncertainties of up to $0.1$, we consider it a very reasonable error.

\subsection{Downstream tasks}

\begin{table*}
\centering
\caption{Accuracies obtained on the validation sets for six classification tasks, trained using five different schemes.}
\label{table:acc}
\begin{tabular}{lccccc}
\toprule
&			from scratch  & \multicolumn{2}{c}{feature extraction} & \multicolumn{2}{c}{fine-tuning} \\ \cmidrule(lr){3-4} \cmidrule(lr){5-6} 
&         			      & ImageNet        & magnitudes      & ImageNet        & magnitudes      \\ \cmidrule(lr){1-6}
EF-15 &   0.4227 \textpm 0.0256 & 0.2642 \textpm 0.0212 & \textbf{0.3271 \textpm 0.0141} & 0.3683 \textpm 0.0222 & \textbf{0.4048 \textpm 0.0245} \\
EF-2  &   0.9683 \textpm 0.0082 & \textbf{0.9697 \textpm 0.0040} & 0.9580 \textpm 0.0056 & \textbf{0.9893 \textpm 0.0016} & 0.9687 \textpm 0.0045 \\
EF-4  &   0.8729 \textpm 0.0153 & \textbf{0.8000 \textpm 0.0271} & 0.7801 \textpm 0.0013 & \textbf{0.8774 \textpm 0.0035} & 0.8322 \textpm 0.0080 \\
MG    &   0.6336 \textpm 0.0000 & 0.7825 \textpm 0.0006 & \textbf{0.8041 \textpm 0.0045} & \textbf{0.9580 \textpm 0.0042} & 0.9360 \textpm 0.0016 \\
SGQ   &   0.8722 \textpm 0.0038 & 0.6624 \textpm 0.0014 & \textbf{0.7255 \textpm 0.0015} & 0.8760 \textpm 0.0011 & \textbf{0.8828 \textpm 0.0008} \\
SG    &   0.9897 \textpm 0.0035 & 0.9366 \textpm 0.0005 & \textbf{0.9799 \textpm 0.0010} & 0.9901 \textpm 0.0006 & \textbf{0.9928 \textpm 0.0004} \\
\toprule
\end{tabular}
\end{table*}

Training and evaluation were carried out according to the setup described in Section~\ref{sec:method_training}, using the datasets described in Section~\ref{sec:method_downstream}. Table~\ref{table:acc} shows validation accuracies for classifiers trained on the five schemes: from scratch, using features extracted from a model pre-trained on ImageNet, using features extracted from a model pre-trained on magnitudes, fine-tuning a model pre-trained on ImageNet and fine-tuning a model pre-trained on magnitudes. Each scheme was run three times to account for randomness in the optimization process. Reported results are the average and standard deviation of the three runs. 

For fine-tuning, models pre-trained on magnitudes perform better by a significant margin for half of the datasets, including both datasets from S-PLUS, but perform worse for most datasets from SDSS.

\begin{figure}[!hb]
    \centering
    \includegraphics[width=0.88\columnwidth]{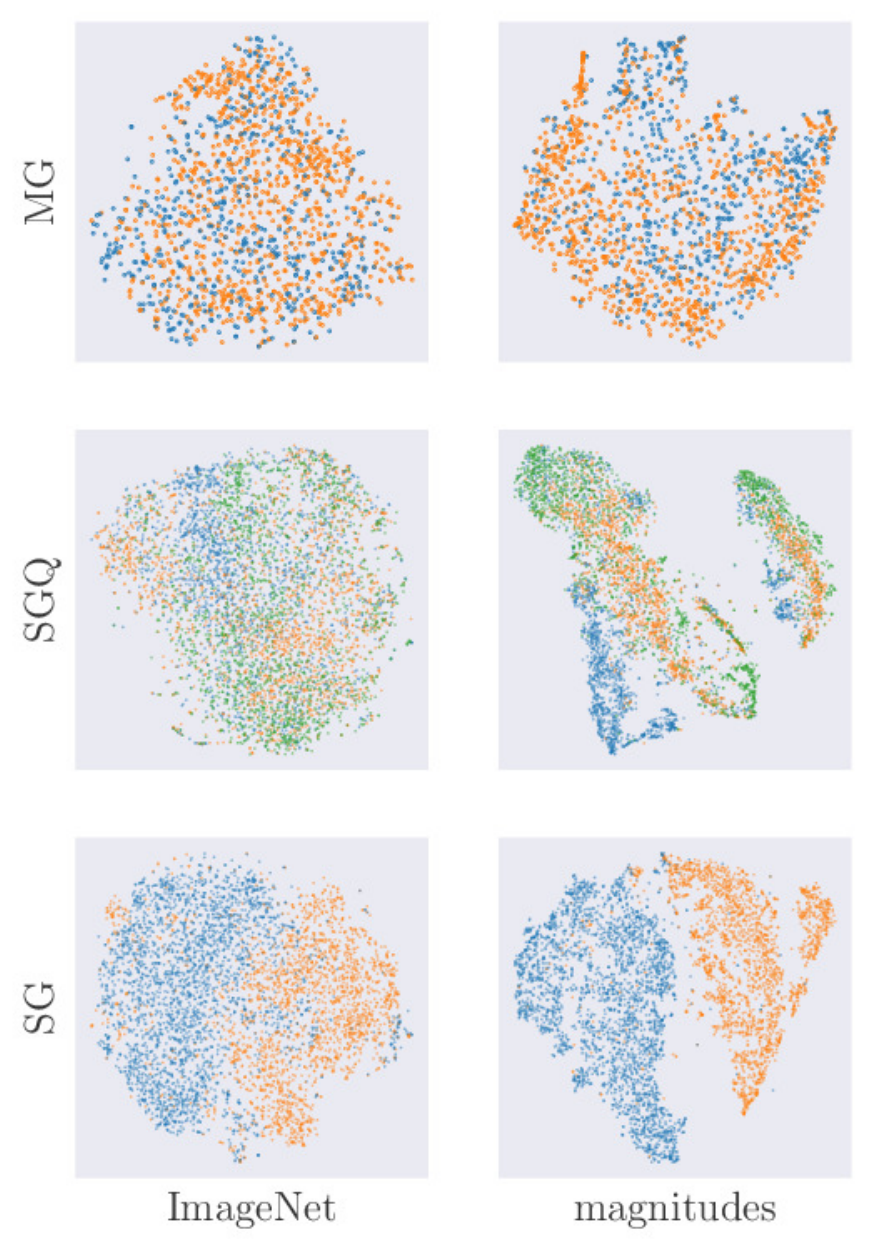}
    \caption{Projections of features extracted from validation sets using models pre-trained on ImageNet and on magnitudes.}
    \label{fig:projections}
\end{figure}

For feature extraction, models pre-trained on magnitudes perform better by large margins for 2/3 of the datasets, including both datasets from S-PLUS and two datasets from SDSS. Figure~\ref{fig:projections} shows two-dimensional projections of features extracted from validation sets of some of the datasets, created with t-SNE~\cite{maaten2008visualizing}, with perplexity=50, where we see that class separation is improved for the features extracted using the model trained on magnitudes, when compared with those extracted using the model trained on ImageNet.

\begin{figure*}[!ht]
    \begin{minipage}[l]{1.0\columnwidth}
        \centering
        \includegraphics[width=\linewidth]{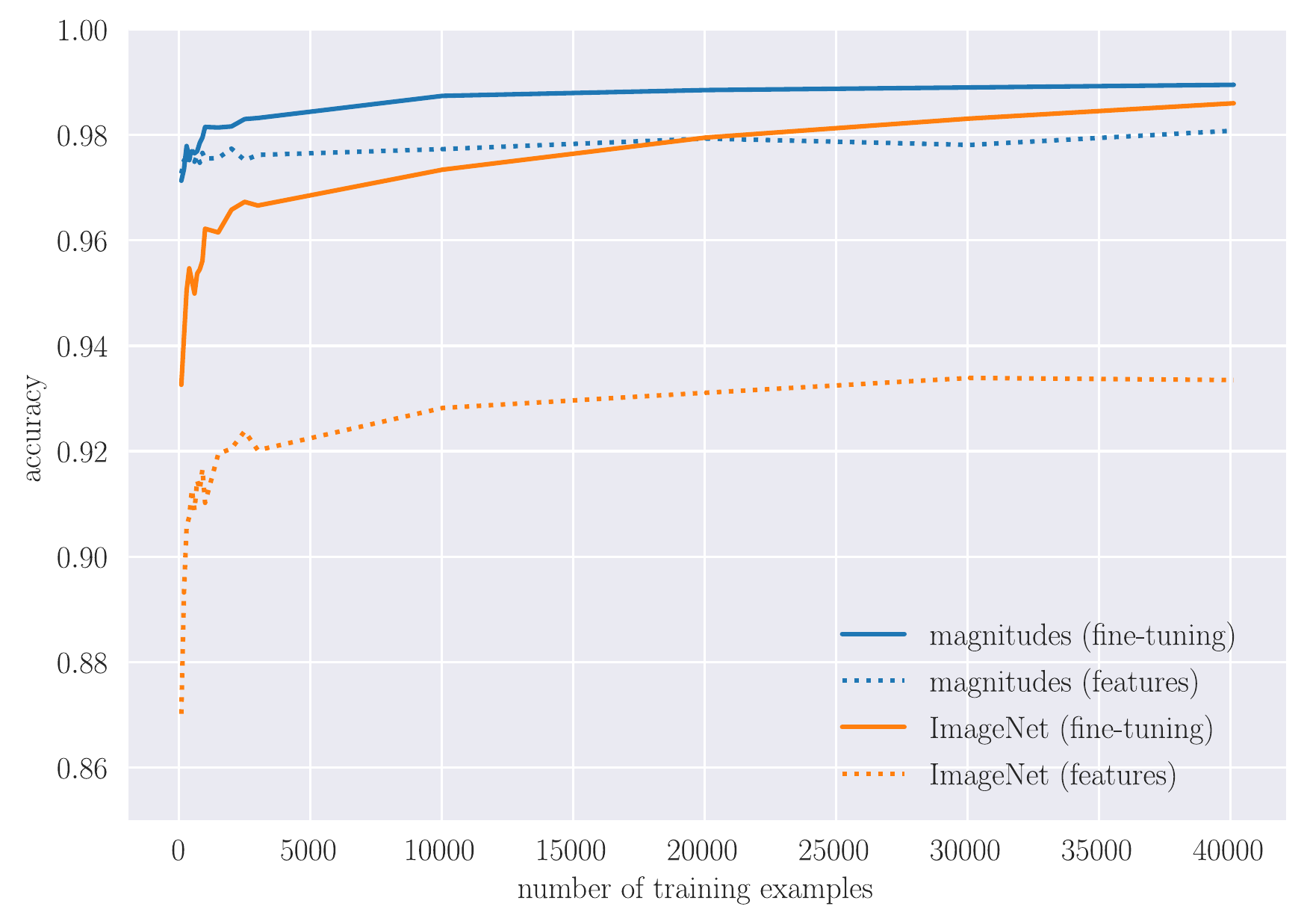}
    \end{minipage}
    \hfill{}
    \begin{minipage}[r]{1.0\columnwidth}
        \centering
        \includegraphics[width=\linewidth]{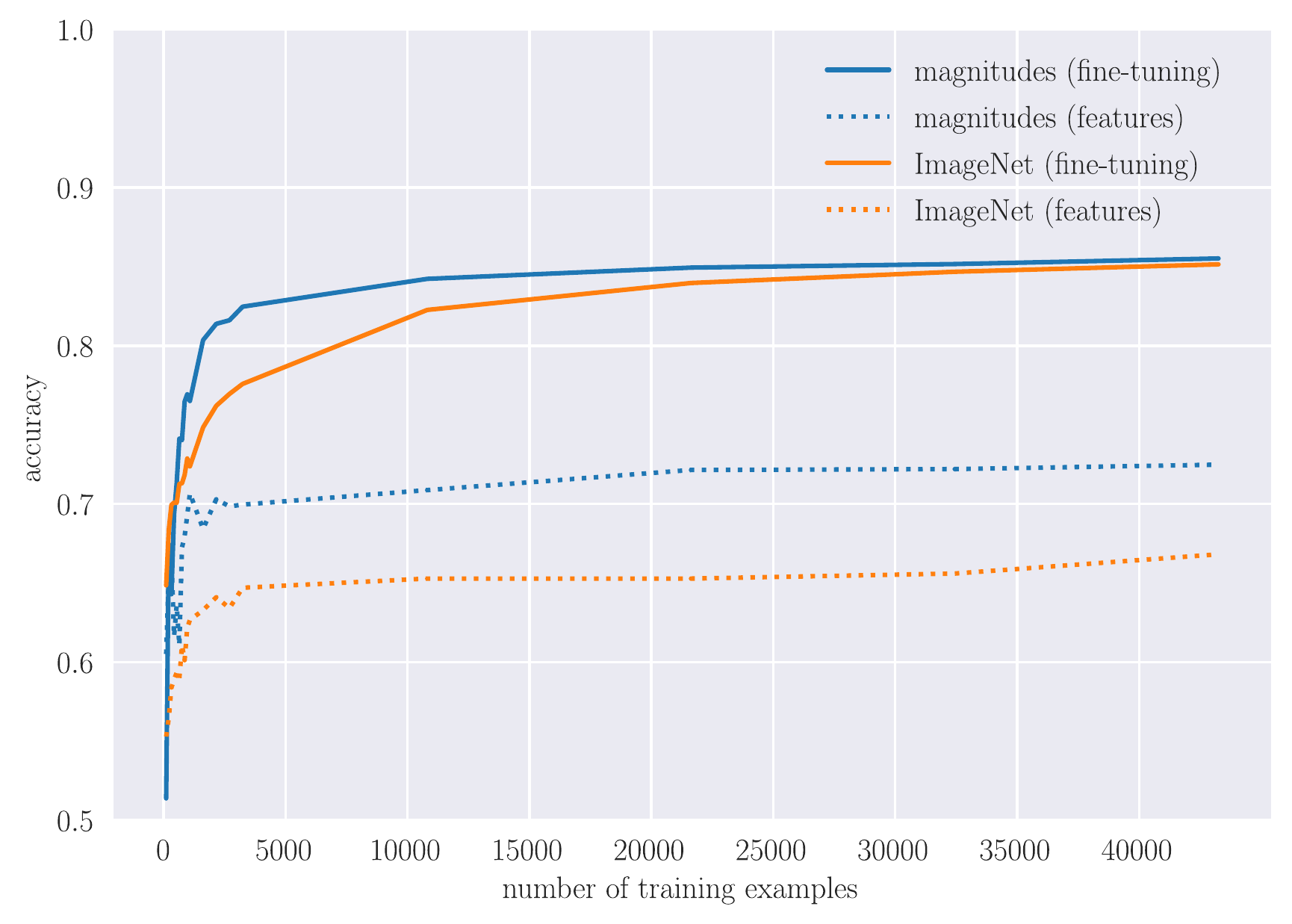}
    \end{minipage}
    \caption{Accuracies on the validation sets of the SG (left) and SGQ (right) datasets as a function of training set size.}
    \label{fig:acc_curves}
\end{figure*}

\subsection{Low-data regime}

Training and evaluation were carried out according to the setup described in Section~\ref{sec:method_low_data}. Figure~\ref{fig:acc_curves} shows accuracy curves as a function of the training set size for the datasets SG and SGQ. Dashed lines refer to classifiers trained on features extracted from the pretext task or from the ImageNet model, and solid lines refers to classifiers trained with fine-tuning from the pretext task or from the ImageNet model.


For feature extraction, the gap between pre-training on magnitudes and on ImageNet remains large and nearly constant regardless of training set size. For fine-tuning, the gap between pre-training on magnitudes and on ImageNet is larger for small training sets and decreases as more training data is added. Table~\ref{table:acc_littledata} shows accuracy values for a training set size of 0.25\% of the total, which corresponds to nearly 100 training examples for each dataset. For such a small training set, it can be seen that pre-training on magnitudes can yield accuracies that are up to 10 pp higher when compared to ImageNet. 

\begin{table}[!ht]
\centering
\caption{Accuracies on the validation sets \newline for a training set of approximately 100 examples.}
\label{table:acc_littledata}
\begin{tabular}{lccc}
\toprule
					&			 &	SG		&	SGQ     \\ \cmidrule(lr){3-4}
                	& ImageNet	 &	0.8702	&	0.5529	\\
feature extraction	& magnitudes &	\textbf{0.9727}	&	\textbf{0.6051}	\\
					& diff.      &	+0.1015	&	+0.052  \\ \cmidrule(lr){2-4}
             		& ImageNet	 &	0.9326	&	\textbf{0.6485}	\\
fine-tuning 		& magnitudes &	\textbf{0.9713}	&	0.5138 	\\
					& diff.		 &	+0.0387	&	-0.1346 \\
\toprule
\end{tabular}
\end{table}

%% file: 050_discussion.tex
We found that it is possible to use astronomical properties of the objects which are easy and cheap to compute as a way to leverage unlabeled images and learn high-quality representations. Representations learned from astronomical properties are compared to representations extracted from a model trained on ImageNet, which is widely regarded as an universal way of extracting image representations via deep learning, regardless of the domain of the images.

Through evaluation using classification tasks and inspection of projections of the representations, we show that learning astronomical properties in a self-supervised manner generates representations that are at least as competitive as representations extracted from a model trained on ImageNet. We consider our approach specially useful in three situations: (i) when there is little labeled data, (ii) when ImageNet weights for the backbone architecture of interest are not easily available, (iii) when performing unsupervised tasks, such as clustering or outlier detection, where domain-specific representations can lead to more easily separable groups.

For the SG and SGQ datasets, representations extracted from the model trained on magnitudes yielded better results. Particularly, we see that with as little as 100 training examples, our approach is capable of achieving significantly greater accuracy when compared with ImageNet, both in fine-tuning and in feature extraction settings. This might enable astronomers to quickly label a few samples and get a working classifier with little effort.

For the other datasets, results are less clear. We believe that this is due to the fact that the other datasets were extracted from SDSS, a sky survey that uses a set of five filters, whereas in our pretext task we train a model to predict magnitudes in twelve passbands. This means that, when predicting twelve magnitudes from SDSS images, we might be trying to extract information that in fact is not present in those images. We believe other easily computed astronomical properties could be explored to help alleviate this problem.

%% file: 060_conclusion.tex
In this work, we propose a self-supervised approach for learning representations from unlabeled astronomical images. We compare the performance of classifiers using learned representations using our approach against ImageNet features, and show that in most cases our approach produces better results and requires much less labeled data to obtain those results.

Possible directions for future work are (i) exploring additional astronomical properties that may be more easily transferrable between different sky surveys, and (ii) evaluating the performance of learned representations on unsupervised tasks such as clustering and outlier detection. This kind of exploratory analysis can yield discovery of fine-grained subclasses, such as different types of stars and galaxies, and of new classes, such as supernova explosions and asteroids.